\documentclass[english]{scrartcl}

\title{Abstract Attribute Exploration\\ with Partial Object
  Descriptions}
\author{Daniel Borchmann \and Bernhard Ganter}
\date{\today}

\usepackage[utf8]{inputenc}
\usepackage[T1]{fontenc}
\usepackage{babel}

\usepackage{etex}
\usepackage{etoolbox}
\usepackage{amsmath,amssymb,mathtools}
\usepackage{latexsym,mathabx,stmaryrd}
\usepackage{graphicx}
\usepackage{enumitem}
\usepackage{array}
\usepackage{booktabs}
\usepackage{microtype}
\usepackage{fixltx2e}
\usepackage{tikz}
\usepackage{csquotes}
\usepackage{fca}
\usepackage[hidelinks]{hyperref}

\usepackage[scale=0.95]{tgpagella}
\usepackage[scale=0.92]{tgheros}
\usepackage[scaled=0.83]{beramono}
\usepackage{mathpazo}

\DeclareMathOperator{\Cn}{Cn}
\DeclareMathOperator{\Th}{Th}
\newcommand{\set}[1]{{\ensuremath{\{\,#1\,\}}}}
\newcommand{\abs}[1]{{\ensuremath{\lvert#1\rvert}}}
\DeclareMathOperator{\Imp}{Imp}

\usepackage[thmmarks,amsmath,hyperref,thref]{ntheorem}

\makeatletter

\newtheoremstyle{standard}%
{\item[\hskip\labelsep \theorem@headerfont ##2\ ##1\theorem@separator]}%
{\item[\hskip\labelsep \theorem@headerfont ##2\ ##1\ (##3)\theorem@separator]}
\theoremstyle{standard}
\theoremheaderfont{\normalfont\bfseries}
\theorembodyfont{\slshape}
\newtheorem{Theorem}     {Theorem} [section]

\newtheorem{Proposition} [Theorem] {Proposition}

\newtheorem{Corollary}   [Theorem] {Corollary}
\theoremheaderfont{\normalfont\bfseries}
\theorembodyfont{\normalfont}
\theoremsymbol{\ensuremath{\diamondsuit}}
\theoremheaderfont{\normalfont\bfseries}
\theorembodyfont{\normalfont}
\theoremsymbol{}

\theoremsymbol{\ensuremath{\diamondsuit}}

\theoremheaderfont{\normalfont\bfseries}
\theorembodyfont{\normalfont}
\theoremsymbol{\ensuremath{\diamondsuit}}
\newtheorem{Definition}  [Theorem] {Definition}
\theoremstyle{nonumberplain}
\theoremheaderfont{\normalfont\itshape}
\theorembodyfont{\normalfont\upshape}
\theoremsymbol{\ensuremath{\Box}}
\newtheorem{Proof}                 {Proof}

\newtheoremstyle{standardreversed}%
{\item[\hskip\labelsep \theorem@headerfont ##1\ ##2\theorem@separator]}%
{\item[\hskip\labelsep \theorem@headerfont ##1\ ##2\ (##3)\theorem@separator]}
\theoremstyle{standardreversed}
\theoremheaderfont{\normalfont\bfseries}
\newtheorem{AlgorithmInternal}{Algorithm}

\makeatother

\def\Neu#1{\textbf{#1}}

\def\calC{{\mathcal C}}
\def\calE{{\mathcal E}}
\def\calL{{\mathcal L}}
\def\background{{\mathcal K}}

\def\domaiN{\mathcal{D}}

\DeclareMathOperator{\PM}{\mathfrak{P}}
\DeclareMathOperator{\clopp}{c}

\usepackage[backend=bibtex,doi=false,url=false]{biblatex}
\usepackage{caption}

\DeclareCaptionFormat{myformat}{#1#2#3\hrulefill}
\usepackage{listings}

\lstdefinelanguage{PseudoCode}{
  alsoletter={-,?,/,*},
  basewidth=0.47em,
  morekeywords=[0]{
    loop,with,recur,let,end,define,if,
    nil,then,else,return,while,do,
    read,and,where,input,output,procedure,
    forever,or,exit,forall,being,each,
    descending,assert,where,choose,returns,add,to
  },
  keywordstyle=[0]\bfseries\ttfamily,
  morekeywords=[1]{explore,attributes,make-exploration-base,query,
    normalize,update-journal,implications,background-knowledge},
  keywordstyle=[1]\upshape\ttfamily,
  classoffset=0,
  numbers=none,
  mathescape=true,
  morecomment=[l]{;},
  commentstyle={\itshape},
}

\usepackage{todonotes}

\begin{document}

\maketitle

\begin{abstract}
  Attribute exploration has been investigated in several studies, with
  particular emphasis on the algorithmic aspects of this knowledge
  acquisition method. In its basic version the method itself is rather
  simple and transparent. But when background knowledge and partially
  described counter-examples are admitted, it gets more
  difficult. Here we discuss this case in an abstract, somewhat
  \enquote{axiomatic} setting, providing a terminology that clarifies
  the abstract strategy of the method rather than its algorithmic
  implementation.
\end{abstract}

\section{Introduction}
\label{sec:introduction}

\textbf{Attribute Exploration} is a fairly well-known and well-tried
knowledge acquisition technique based on Formal Concept
Analysis~\cite{fca-book}. It is similar to \textbf{Query Learning}, as
described in~\autocite{journals/ml/AngluinFP92}.  Though quite simple
in its basic idea, it can become rather intricate in its many
extensions and modifications (see Ganter \& Obiedkov (in preparation)
for an overview). Readers who are unfamiliar with this method may find
motivating examples in the literature.

The aim of this paper is not to add yet another generalization, but to
present an abstract version of one of the more basic variants, namely
attribute exploration with \emph{partial object
  descriptions}~\cite{conf/ijcai/BaaderGSS07}.  For this we shall make
use of the formalization of a \emph{domain expert}, as developed by
the first author~\autocite{conf/icfca/Borchmann15}.  We thereby hope
to make the method more transparent. It is our impression that such a
strictly formal setting might eventually be more practical.

In general, the purpose of an attribute exploration is to classify, by
querying an expert, all admissible attribute combinations of a given
attribute set. A typical query consists of a logical proposition,
usually an implication, that is presented to the domain expert. The
expert has to decide if that proposition holds for all admissible
sets. In case that it does not, the expert is asked to provide a
counter-example, i.e., an admissible set that is not a model of the
proposition. In the version discussed here we allow for partially
specified counter-examples.
 
The set of models of an arbitrary set of implications is closed under
intersections. Therefore, if only implications are used, admissible
sets can only be classified up to intersections.

\section{The parts of the exploration method}

The attribute exploration method consists of four modules which have
the following purposes:
\begin{description}
\item [The exploration schema ] specifies the framework of an
  exploration. It defines the universe which is to be explored.
\item [The exploration base ] contains the current status of an
  exploration.
\item [The domain expert ] is the reliable information source for an
  exploration.
\item [The exploration engine] consists of the algorithmic machinery
  for the communication between domain expert and exploration base.
\end{description}

These four parts will now be described more precisely. Before doing
so, we shortly sketch the nature of these four modules.

The exploration schema specifies the set of all attribute combinations
under investigation (admissible or not). We call this the universe of
the respective exploration. Moreover, the schema may contain
additional background knowledge.

The exploration base has three parts. One contains the
counter-examples that have already been accepted or inferred. The
second contains the implications which have been validated or
inferred.  Both lists can be modified by the domain expert and by the
exploration engine. A journal file, being the third part of the base,
records all such modification.

The domain expert receives questions in the form of implications. Each
asked implication is either validated or disproved by giving a
(partially specified) counter-example. 
 
The exploration engine examines at any point of an exploration the
current exploration base if there exists an implication that holds for
all collected counter-examples, but which cannot be inferred from the
validated implications. If so, then such an implication is presented
to the domain expert. Otherwise, the exploration terminates. The
exploration engine also modifies the exploration base according to
(valid) inference rules.

\section{The exploration schema}

A (finite) set $M$ of attributes under investigation is specified. 
The \textbf{universe} is the power set $(\PM(M),\subseteq)$. 

The set to be determined by the exploration is some subset
$\calE\subseteq\PM(M)$, which however is not directly accessible, but
only through questioning some \enquote{domain expert}. The subsets
which are elements of $\calE$ are called \textbf{admissible sets} or,
synonymously, \textbf{counter-examples}. The questions to the domain
expert are formulated as \textbf{implications} $R\to S$, where $R$ and
$S$ are subsets of $M$. A set $X\subseteq M$ \textbf{respects} (or is
a \textbf{model} of) an implication $R\to S$, if $S\subseteq X$ or
$R\not\subseteq X$.  Otherwise, if $R\subseteq X$ and $S\not\subseteq
X$. then $X$ \textbf{refutes} the implication $R\to S$. By validating
an implication $R\to S$ the domain expert states that this implication
is respected by all admissible sets. In other words, if an admissible
set contains all elements of $R$, then it also must contain all
elements of $S$. If an implication $R\to S$ is not confirmed, then the
expert provides a counter-example, i.e., some set $E\in\calE$ that
refutes the implication $R\to S$. We allow, however, that the domain
expert does not specify such a counter-example precisely. Instead, a
partial description may be given in form of a pair $(U,V)$ of sets
$U\subseteq V\subseteq M$ such that $R\subseteq U$ and $S\not\subseteq
V$. By providing such a pair $(U,V)$ the domain expert expresses that
there is some admissible set $E$ with $U\subseteq E\subseteq V$ (which
then necessarily refutes the implication $R\to S$). Such an $E$ is
called an \textbf{admissible completion} of the \textbf{partial
  counter-example} $(U,V)$.

Some \textbf{background knowledge} may be present that rules out parts
of the universe as non-admissible, so that only some subset
$\background\subseteq\PM(M)$ of the power set needs to be
investigated. Such background knowledge usually is given in the form
of propositional formulas over $M$. It excludes all subsets that are
not models of these formulas.  One possible data type for such
background formulae is that of set families $(A,B_1,B_2,\ldots,B_k)$,
representing \textit{cumulated clauses}
\begin{equation*}
\bigwedge A\to \bigvee_{i=1}^k\bigwedge B_i,
\end{equation*}
where $A$ as well as $B_1,\ldots,B_k$ are subsets of $M$. The
condition expressed by such a formula is that any admissible set which
contains all the attributes from $A$ must also contain all attributes
from $B_i$, for at least one $i$. There are algorithms for the
exploration engine that use such cumulated clauses.

If the background knowledge excludes the top element $M$, i.e., if the
combination of all attributes is not admissible, then we must allow
for \emph{indefinite} implications. In this case, the top 
element is denoted as $\bot$, and $A\to \bot$ indicates that $A$ is
not contained in any admissible set.  

We present here what we call the \emph{set-based exploration
  schema}. More abstract versions are under discussion, but have not
yet been worked out.

\section{The exploration base}
An exploration base consists of three parts: a journal file, the
collection of accepted counter-examples, and the collection of
validated implications. These sets may be empty in the beginning of an
exploration. 

Both collections are subsets of the order relation $\subseteq$, which
means that each entry of these collections is a pair $(R,S)$ of
elements from $\PM(M)$ with $R\subseteq S$. How do such pairs
represent counter-examples and implications?

A pair $(R,S)$ of subsets represents the implication $R\to S$, which
is considered to hold true iff $R\subseteq A$ implies $S\subseteq A$ for 
all admissible sets $A\subseteq M$. W.l.o.g.\ we may assume that $R\subseteq
S$, because  $R\to S$ is logically equivalent to $R\to R\cup S$, since
both implications are respected by the same sets. 

A partial counter-example is a pair $(U,V)$ of subsets with
$U\subseteq V\subseteq M$. It refutes the implication $R\to S$ iff
\begin{equation*}
R\subseteq 
U\quad\mbox{and}\quad S\not\subseteq V.
\end{equation*}
It is assumed that the domain expert only provides partial
counter-examples which have an admissible completion. That implies
that throughout each exploration process the exploration base remains
\textbf{consistent} in the sense that it has at least one
\textbf{realizer}, i.e., that all partial examples can be completed in
such a way that they respect all validated implications and are not
excluded by the background knowledge.

We shall prove later that an exploration base can be checked for
consistency without knowing what the admissible sets are. 
We warn the reader that for an exploration base to be consistent, it
is not sufficient that no counter-example refutes a validated
implication.

We write $\calL$ for the list of implications and $\calC$ for the list
of partial counter-examples in the base. We ignore the journal file
for the time being and abbreviate the exploration base as
$(\calL,\calC)$. Such a base may have more than one realizer, but it
has --if consistent-- always a largest one. That consists of all
$\calL$-closed sets which are compatible with the background
knowledge. 

Our expectation is that during an exploration process the exploration
base gets enriched so that it eventually converges to a specific
realizer, representing the domain that is explored. To make this more
precise, consider two exploration bases $(\calL_1,\calC_1)$ and
$(\calL_2,\calC_2)$ for the same schema. We say that
$(\calL_1,\calC_1)$ is more (or equally) \Neu{expressive} than
$(\calL_2,\calC_2)$, if every realizer of  $(\calL_1,\calC_1)$ also is
a realizer of  $(\calL_2,\calC_2)$. Easier to check is the following
condition: We call  $(\calL_1,\calC_1)$ is better (or equally)
\Neu{focused}  than $(\calL_2,\calC_2)$, if the following two
conditions are fulfilled:
\begin{enumerate}
\item For each implication $A\to B$ in $\calL_2$ there is an
  implication $C\to D$ in $\calL_1$ with
  \begin{equation*}
    C\subseteq A\subseteq B\subseteq D.
  \end{equation*}
\item For each partial counter-example $(U,V)$ in $\calC_2$ there is a
  partial counter-example $(X,Y)$ in $\calC_1$ with
  \begin{equation*}
    U\subseteq X\subseteq Y\subseteq V.
  \end{equation*}
\end{enumerate}
It is easy to see that a better focused base automatically is more or
equally expressive.

The exploration base is modified by the domain expert and by the
exploration engine. In a later stage of an exploration procedure it
might not at all be obvious why a certain entry in the base is
present.  We therefore find it advisable to keep track of all
modifications. That is what the journal file is for.

\section{The domain expert}
\label{sec:domain-expert}

The domain expert is the unique information source for the
exploration. If this expert is a human, a group, or an algorithm, does
not really matter for our considerations. Essential is the (probably
unrealistic) assumption that the expert is consistent in the sense
that the given answers are compatible with a realizer. This is now
formalized. 

Consider a mapping $p$ that maps implications to counter-examples or
to a symbol indication that the implication is true. Recall that both
implications and counter-examples are comparable pairs of subsets of
$M$, and recall that the subset-order relation formally is the set of all
comparable pairs,
\begin{equation*}
  \subseteq\; = \{(A,B)\mid A,B\in \PM(M), A\subseteq B\}.
\end{equation*}
Thus such a mapping is of the form
\begin{equation*}
  p\colon\;\subseteq\;\to\;\subseteq\cup\;\{\top\},\quad\mbox{where }
  \top\notin\;\subseteq.
\end{equation*}
For better readability we shall write $R\to S$ for implications
and $(U,V)$ for partially specified counter-examples. We call
$p(\subseteq)\setminus\{\top\}$ the set of partial
\Neu{counter-examples of} $p$ and $\Th(p):=p^{-1}(\top)$ the
\Neu{theory} of $p$. 
\begin{Definition}\label{def:domain-expert}%
A mapping $p:\;\subseteq\;\to\;\subseteq\cup\;\{\top\}$ is called a
\Neu{domain expert} (on the universe $(\PM(M),\subseteq)$, with
background knowledge $\background\subseteq \PM(M)$) iff it satisfies
the following conditions:  
  \begin{enumerate}[label=\textit{\roman{*}}.]
  \item\label{item:4}%
    \emph{$p$ gives partial counter-examples to false implications}: if
    $p(X\to Y) = (T,U) \neq \top$, then $(T,U)$ is a
    partial counter-example for $X \to Y$, i.e., $X \subseteq T$ and
    $Y \not\subseteq U$.
    
  \item\label{item:3}%
    \emph{$p$ does not give partial counter-examples to valid
      implications}: if $X \to Y, V \to W$ are such that
    \begin{equation*}
      p(X \to Y) = \top\quad\mbox{and}\quad p(V \to W) = (T, U),
    \end{equation*}
    then $(T, U)$ is not a partial counterexample for $X \to Y$, i.e., $X
    \subseteq T$ implies $Y\subseteq U$.

  \item\label{item:1}%
    \emph{$p$ is consistent with the background knowledge}: Each
    counter-example $(T,U)$ of $p$ has an \Neu{compatible completion},
    by which we mean an element $D\in \background{}$ with $T\subseteq
    D\subseteq U$, that respects all implications $R\to S$ with
    $p(R\to S)=\top$.  
  \end{enumerate}
\end{Definition}
Domain experts are meant to \enquote{represent domains}.  What we
mean by this is formalized in the following definition.

\begin{Definition}\label{def:domain}
  Let $p$ be a domain expert on $(\PM(M),\subseteq)$ with background knowledge
  $\background{}\subseteq\PM(M)$.  A \emph{domain} for $p$ is a set
  $\domaiN \subseteq S$ such that we have
  \begin{itemize}
  \item $p(X \to Y) = \top \iff \mbox{all }D\in\domaiN \mbox{ respect }
    X \to Y$, and  
  \item each partial counter-example $(T,U)$ of $p$ has a compatible
    completion $D \in \domaiN$.
  \end{itemize}
  We say that $p$ \emph{represents a domain} if there is a domain for $p$.
\end{Definition}

\begin{Proposition}\label{prop:represents}
  Let $p$ be a domain expert on $(\PM(M),\subseteq)$ with background
  knowledge $\background{}\subseteq \PM(M)$.  Then $p$ represents a domain.
\end{Proposition}
\begin{Proof}
For each partial counter-example $(T, U)$ of $p$ let $D_{(T,U)}\in\background{}$  
be a compatible completion.  Such completions exist because
  $p$ is consistent with the background knowledge $\background{}$.  Define 
  \begin{equation*}
    \domaiN \coloneqq \set{ D_{(T, U)} \mid (T, U) \mbox{ a partial
        counter-example of }p}.
  \end{equation*}
Obviously, $D$ fulfills the second condition of
Definition~\ref{def:domain}. It remains to verify the first condition.
So let $X\to Y$ be an implication which is respected by all
$D\in\domaiN$. In order to show that $p(X\to Y)=\top$, we assume
the contrary, that $p(X\to Y)=(T,U)$ for some partial counter-example
$(T,U)$. If $X\subseteq T$, then $Y\subseteq D_{(T,U)}\subseteq U$, 
since $D_{(T,U)}$ respects $X\to Y$. But that contradicts the first
condition of Definition~\ref{def:domain-expert}, and proves that
$p(X\to Y)=\top$. 
Finally, assume that  $p(X\to Y)=\top$, and let $D\in\domaiN$. We
must show that $D$ respects $X\to Y$. But that is immediate from the
fact that $D=D_{(T,U)}$ is a compatible completion of some
counter-example $(T,U)$.
\end{Proof}

Conversely, when we are given a set $\domaiN\subseteq S$, then we
can easily obtain a domain expert $p_{\domaiN}$ that represents
the domain $\domaiN$.  We can do this by defining
\begin{equation*}
  p_{\domaiN}(X\to Y) \coloneqq
  \begin{cases}
    \top & \text{if all elements of } \domaiN \text{ respect } X\to Y,\\
    (X, D) & \text{otherwise},
  \end{cases}
\end{equation*}
where $D$ is some element of $\domaiN$ that does not respect $X \to Y$.
\bigbreak

Proposition~\ref{prop:represents} sheds some light on the difference
between \enquote{admissible} and \enquote{compatible}
completions. According to this proposition the requirements of
Definition~\ref{def:domain} describe the behavior of a domain expert
for \emph{some} domain, but not necessarily for the system $\calE$ of
admissible sets that we plan to explore.

Proposition~\ref{prop:represents} also shows that the exploration base
remains consistent as long as its implications and partial examples
come from a domain expert $p$. Obviously a domain for $p$ is a
realizer, even if the exploration is still incomplete.

If $p$ is a domain expert and $X\subseteq\PM(M)$ is arbitrary, then there is
always a largest subset $Y$ of $M$ such that $p(X,Y)=\top$.  This can
be seen as follows: let $\domaiN$ be a domain for $p$ and let%
\begin{equation*}
 \clopp_{p}(X) := \bigwedge\{D\in\domaiN\mid X\subseteq D\}.
\end{equation*}
The implication $X\to \clopp_{p}(X)$ is respected by all $D\in\domaiN$, but
for any larger element $Z\not\subseteq \clopp_{p}(X)$ there is some
$D\in\domaiN$ with $X\subseteq N$ and $Z\not\subseteq D$, refuting the
implication $X\to Z$.

Note that $\clopp_{p}(X)$ is not necessarily admissible. If
$\{D\in\domaiN\mid X\le D\}$ is empty, then $\clopp_{p}(X)$ is $M$.
\bigbreak

Technically, the condition that domain experts do not give
counter-examples to valid implications is not necessary.  Suppose that
we are given a mapping $p$ as above that gives counter-examples to
false implications and is consistent with the background knowledge.
We can define a domain expert $q$ on $(\PM(M),\subseteq)$ such that
$\Th(p)=\Th(q)$ as follows: 
\begin{equation*}
  q(X \to Y) \coloneqq
  \begin{cases}
    \top & \text{if } p(X \to Y) = \top, \\
    (\Th(p)(T), U) & \text{if } p(X \to Y) = (T, U).
  \end{cases}
\end{equation*}
Then $\Th(p) = \Th(q)$.  Each partial counter-example  $(T, U)$ of $p$ for
$X \to Y$ has a compatible completion $D$ satisfying
\begin{equation*}
  T \le \Th(p)(T) \le D \le U.
\end{equation*}
Therefore also $(\Th(p)(T), U)$ is a counter-example to $X \to Y$.
Additionally $(\Th(p)(T), U)$ cannot be a counterexample to any
implication in $\Th(q) = \Th(p)$, and $q$ is consistent with
$\background{}$.  Thus $q$ is a domain expert on $(\PM(M),\subseteq)$
such that $\Th(p) = \Th(q)$.

On the other hand, the condition that a domain expert does not give
counterexamples to valid implications is very intuitive and we shall
keep it for this reason.
\bigbreak

Condition~\emph{iii} of Definition~\ref{def:domain-expert} seems more
problematic, because it is hard to verify, actually is infeasible.
This is because checking for a pair $(T, U)$ of subsets 
$T \subseteq U \subseteq M$ whether there exists a model $N$ of
the background knowledge between $T$ and $U$ is NP-complete: when choosing
$T = \emptyset$ and $U = M$, solving this problem amounts to deciding
whether the background knowledge is satisfiable.

From a practical point of view, however, this observation does not
necessarily mean much of a problem, for two reasons:
\begin{enumerate}
\item We may rely on the efficiency of modern high-performance
  SAT-solvers. Condition~\emph{iii} can easily be transformed into a
  propositional satisfiability problem, at least if the background
  knowledge is given in propositional form. What must be checked is if
  the background knowledge together with all validated implications 
  has a model between  $T$ and $U$.
\item In practice, the formulae for the background knowledge are
  usually quite simple. This is due to the fact that often they
  describe relations between the attributes as a result of
  \emph{conceptual scaling}. It has been shown that a complexity
  measure for background knowledge can be introduced such that
  testing  Condition~\emph{iii} for background knowledge of bounded
  complexity can be solved in polynomial time.
\end{enumerate}

\begin{Corollary}
  For every domain expert $p$ the set $\Th(p)$ is closed under
  inference.
\end{Corollary}
\begin{Proof}
  By Proposition~\ref{prop:represents}, the expert $p$ represents some
  domain $\mathcal{D}$.  Then
  \begin{equation*}
    p(X \to Y) = \top \iff \mathcal{D} \models (X \to Y).
  \end{equation*}
  Let $\mathcal{L} \cup \set{ X \to Y } \subseteq \Imp(M)$ such that
  $\mathcal{L} \models (X \to Y)$ and $\mathcal{L} \subseteq \Th(p)$.
  Then $\mathcal{D} \models \mathcal{L}$ and therefore $\mathcal{D}
  \models (X \to Y)$.  This means $p(X \to Y) = \top$ and thus $(X \to
  Y) \in \Th(p)$ as required.
\end{Proof}

\section{The exploration engine}
Once again, this has several parts.

One is the \textbf{query engine}. It checks if the exploration base is
complete in the sense that all implications which are not refuted by
some counter-example can be inferred from the validated implications. If not, it
generates a question which the domain expert is asked. Such a question
always is an implication which is not refuted by any of the accepted
counter-examples. However, it may happen that an implication is asked
which would make the exploration base inconsistent when added. In such
a case, the domain expert is forced to provide a counter-example.  (To
keep things simple, we do not discuss here the possibility of
\emph{postponing} questions.) How the query engine works is easier to
explain for the case that the exploration base is \emph{normalized}.

The \textbf{normalization engine} attempts to simplify the
counter-examples of the exploration base without changing the
information they contain.
\begin{itemize}
\item Whenever in the base there are a partial counter-example $(U,V)$
  and an   implication $R\to S$ such that $R\subseteq U$, then the
  partial counter-example is replaced by $(U\cup S, V)$.

 (Note that $U\cup S\subseteq V$ is implied by the consistency of the
  exploration base.) 
\item Whenever in the base there are a partial counter-example
  $(U,V)$, an element $v\in V\setminus U$, and an implication
  $U\cup\{v\}\to S$,   where $S\not\subseteq V$, then the example
  $(U,V)$ is replaced by   $(U,V\setminus\{v\})$.
\item Whenever in the base there are counter-examples $(U_1,V_1)$ and
  $(U_2,V_2)$ such that $U_1\subseteq U_2\subseteq V_2\subseteq V_1$,
  then $(U_1,V_1)$ is removed from the base.
\end{itemize}

Observe that for all three modifications a partial counter-example
$(U_1,V_1)$ is replaced by a tighter one $(U_2,V_2)$, i.e., one with
$U_1\subseteq U_2\subseteq V_2\subseteq V_1$. And this is done in a
way that necessarily each compatible completion lies in the tighter
interval. Therefore normalization may make the base better focused,
but does not change its expressiveness.  

There also is the \textbf{inference engine} which decides, for any
suggested implication $A\to B$, if it follows from the implications in
the exploration base together with the background knowledge. The
inference rules are the standard ones of propositional logic. The
inference engine also can compute, for any given set $A\subseteq M$,
the largest set $\calL(A)$ such that the implication $A\to\calL(A)$
follows. Moreover, the inference engine may be used to streamline the
implicational part of the exploration base by removing redundant
implications. Again, the expressiveness of the base remains invariant.

Some complexity considerations are advisable here, since implication
inference under arbitrary propositional background knowledge is again
an $\mathcal{NP}$-complete problem. Solutions similar to the ones
discussed in the previous section are possible, but a simple
alternative is to disregard the background knowledge when computing
the implication inference (which then is very easy). This does not
produce false implications, but may fail to find certain valid
ones. As a result, questions may be asked to the domain expert which
in principle could be answered by the algorithm. This strategy may be
limited to time-critical situations. 

On the other hand, if the size of background knowledge that is
formulated as cumulated clauses is rather small, then implicational
inference is again easy: let $\mathcal{L} = \mathcal{N} \cup
\mathcal{H}$ be a set of clauses such that $\mathcal{N}$ consists of
all cumulated clauses in $\mathcal{L}$ and $\mathcal{H}$ consists of
all of implications in $\mathcal{L}$.  Then assuming $\mathcal{N}$ is
fixed, deciding whether an implication follows from $\mathcal{L}$ can
be done in $\mathcal{O}(\abs{\mathcal{H}} \cdot \abs{M}^{2})$
steps~\cite{journals/dam/GanterK05}.

We come back to the \textbf{query engine}, now for a normalized
exploration base. For any set $A\subseteq M$, define
\begin{equation*}
  A^{+?}:=\bigcap\{V\mid (U,V)\mbox{ is a counter-example in the base
    such that }   A\subseteq U\}.
\end{equation*}

Note that $A^{+?}$ is the largest set for which the implication
\begin{equation*}
  A\to A^{+?}
\end{equation*}
is not refuted by any partial counter-example from the base.

The query engine searches for a set $A\subseteq M$ such that
\begin{equation*}
A=\calL(A)\ne A^{+?}.
\end{equation*}
If such a set is found (preferably a simple one), then the question
\begin{equation*}
A\to A^{+?}
\end{equation*}
is proposed to the domain expert. If no such set exists, then the
exploration is complete. This is stated in the following theorem.
\begin{Theorem}
  If $A=A^{+?}$ holds for all sets that satisfy $A=\calL(A)$, then the
  implications which are not refuted by partial counter-examples are
  precisely those that follow from implications validated by the
  domain expert.
\end{Theorem}
\begin{Proof}
  The implications in the exploration base, together with their
  logical consequences,  are precisely those that follow from the ones
  validated by the expert, since all   modifications were made
  according to valid inferences. So we must show that these are the
  ones which are not refuted by any partial counter-example. Since the
  exploration base remains consistent during an exploration process,
  all such implications are respected by all partial counter-examples.    
  Could there be an implication $A\to B$ which is respected by all
  partial counter-examples, but which does not follow from the ones in
  the base? Since $A\to\calL(A)$ follows from the base, this requires
  $B\ne\calL(A)$, and we may assume $A=\calL(A)$, because $\calL(A)\to
  B$ must also be respected by all partial counter-examples. Moreover
  we have $B\subseteq A^{+?}$, since $A^{+?}$ is the largest set for
  which the implication $A\to A^{+?}$ is not refuted by any partial
  counter-example from the base. Putting the pieces together yields
  $A=\calL(A)\ne B\subseteq A^{+?}$, and therefore $A=\calL(A)\ne A^{+?}$.
\end{Proof}

\section{Attribute exploration}

An attribute exploration starts with specifying an exploration schema
and an initial exploration base. The latter may be empty.  Then the
query engine computes a question, which the domain expert answers. The
answer is added to the exploration base, which then is
normalized. This question-answer process is repeated until the query
engine finds that the exploration base is complete.
Figure~\ref{fig:exploration-algorithm-pseudo-code} show a pseudo-code
implementation of this algorithm.

\begin{figure}[t]
\begin{lstlisting}
define explore($S$, $p$, $\mathcal{C}$)
  assert $S \text{ is an exploration schema}$
  assert $p \text{ is a domain expert for schema } S$
  assert $\mathcal{C} \text{ is a set of counter-examples for schema } S$

  let $\background$ := background-knowledge($S$)
      $M$ := attributes($S$)
      base := make-exploration-base($\mathcal{C}$, $\background$) do

    while query(base) returns implication $A \to B$ do
      if $p(A \to B) = \top$ then
        add $A \to B$ to base
      else
        $(U, V)$ := $p(A \to B)$
        add $(U, V)$ to base
      end if

      normalize(base)
      update-journal(base)
    end while

    return implications(base)
  end let
end define
\end{lstlisting}
  \caption{Abstract Attribute Exploration Algorithm}
  \label{fig:exploration-algorithm-pseudo-code}
\end{figure}

The following claim is now obvious.

\begin{Theorem}
  Let $S$ be an exploration schema on some attribute set $M$ with
  background knowledge $\background$.  Let $p$ be a domain expert on
  $M$, $\mathcal{C}$ be a set of models of $\Th(p)$, and $\background
  \subseteq \Th(p)$.  Then
  \begin{equation*}
    \mathcal{L} \coloneqq \mathsf{explore}(S,p,\mathcal{C})
  \end{equation*}
  is a base of $\Th(p)$ for background knowledge $\background$, i.e.,
  \begin{equation*}
    \Cn(\mathcal{L} \cup \background) = \Th(p).
  \end{equation*}
\end{Theorem}

For the special case that the background knowledge itself consists of
implications (including the case of empty background knowledge) it is
known that the implications validated by the domain expert constitute
the so-called \emph{canonical base}, and that this base is of minimal
possible size. The domain expert thus has to validate as few
implications as possible. It is also known that no similar result hold
for the number of (partial) counter-examples: later examples may make
previously given ones dispensable. Indeed, the normalization procedure
for the base includes canceling partial counter-examples that contain
tighter ones. An optimality result for the number of counter-examples
thus is not to be expected. However, there is a unique minimal
realizer, consisting of the $\bigcap$-irreducible models of
$\Th(p)\cup\background$. Therefore we get the exact number of
indispensable counter-examples \emph{a posteriori}. We know of no
strategy for the domain expert that would lead to irreducible
counter-examples only.

\section{Outlook}
\label{sec:outlook}

The abstract formulation of attribute exploration as we have presented
it in this work does not capture all properties of the classical
algorithm.  The main example, namely that classical attribute
exploration computes the canonical base in the presence of
implicational background knowledge, was briefly mentioned in the the
previous section.  As expert interaction can be assumed to be
expensive, minimizing the number of questions asked to the expert
seems inevitable for the practicability of exploration itself.

One of the logical next steps in the investigation of an abstract
formulation of attribute exploration is thus to examine properties of
the exploration engine and the domain expert which are required to
minimize the number of expert interactions.  The classical algorithm
already gives a possibility to implement the exploration engine such
that expert interaction is minimized.  In this implementation, the
implications returned by the engine are enumerated in a
$\subseteq$-extending order.  One could now ask whether this condition
is really necessary, or even sufficient for minimal expert
interaction.  Recent results on the parallelizability of the
computation of the canonical base seem to suggest that, under some
additional mild assumptions, the latter might indeed be
true~\cite{conf/cla/KriegelB15}.

Another interesting perspective for future research is the comparison
of attribute exploration with the query learning algorithm for
propositional Horn logic from~\cite{journals/ml/AngluinFP92}.  It is
not completely obvious that those two algorithm indeed accomplish the
same.  However, it is known that the algorithm
from~\cite{journals/ml/AngluinFP92} computes the canonical base not
matter what the form of the particular choice of the examples given by
the involved oracles~\cite{journals/ml/AriasB11}.  This is very
similar to the classical attribute exploration algorithm with complete
counterexamples, where the particular choice of the counterexamples
given by the expert does not have an impact on the form of the
computed base.  This suggest that there might be a deeper connection
between attribute exploration and query learning, and it may be worth
discovering this connection for the benefit of both approaches.

\appendix

\printbibliography{}

\end{document}